\pdfoutput=1

\documentclass[11pt]{article}

\usepackage{acl}

\usepackage{times}
\usepackage{latexsym}
\usepackage{amsmath}
\usepackage{bm}
\usepackage{amsfonts}
\usepackage{amssymb}
\usepackage{booktabs}
\usepackage{subcaption}
\usepackage{caption}
\usepackage{graphicx}
\usepackage[T1]{fontenc}

\usepackage[utf8]{inputenc}

\usepackage{microtype}
\usepackage{arydshln}
\usepackage{mathabx}
\usepackage{fontawesome}

\newcommand\scalemath[2]{\scalebox{#1}{\mbox{\ensuremath{\displaystyle #2}}}}

\newcommand{\argmax}{\operatornamewithlimits{argmax}}

\title{Modeling Label Correlations for Ultra-Fine Entity Typing with \\ Neural Pairwise Conditional Random Field}

\makeatletter
\def\@fnsymbol#1{\ensuremath{\ifcase#1\or \dagger\or *\or \ddagger\or
   \mathsection\or \mathparagraph\or \|\or **\or \dagger\dagger
   \or \ddagger\ddagger \else\@ctrerr\fi}}
\makeatother

\newcommand{\damo}{\textsuperscript{\faStarO}}

\newcommand{\code}{\url{http://github.com/modelscope/adaseq/tree/master/examples/NPCRF}}

\author{
    \textbf{Chengyue Jiang}\thanks{$~~$This work was done during Chengyue Jiang's internship at DAMO Academy, Alibaba Group. } ,
    \textbf{Yong Jiang}\damo$^{\ast}$,
    \textbf{Weiqi Wu},
    \textbf{Pengjun Xie}\damo,
    \textbf{Kewei Tu}\thanks{$~~$Yong Jiang and Kewei Tu are corresponding authors.} \\
    \damo DAMO Academy, Alibaba Group, China \\
    \texttt{\{jiangchy1997,kewei.tu.nlp\}@gmail.com;}
    \texttt{wuwq1022@foxmail.com;}\\
    \texttt{\{yongjiang.jy,chengchen.xpj\}@alibaba-inc.com }
}

\begin{document}
\maketitle

\begin{abstract}
    Ultra-fine entity typing (UFET) aims to predict a wide range of type phrases that correctly describe the categories of a given entity mention in a sentence. 
    Most recent works infer each entity type independently, ignoring the correlations between types, e.g., when an entity is inferred as a {\it president}, it should also be a {\it politician} and a {\it leader}.
    To this end, we use an undirected graphical model called pairwise conditional random field (PCRF) to formulate the UFET problem, in which the type variables are not only unarily influenced by the input but also pairwisely relate to all the other type variables.
    We use various modern backbones for entity typing to compute unary potentials, and derive pairwise potentials from type phrase representations that both capture prior semantic information and facilitate accelerated inference. We use mean-field variational inference for efficient type inference on very large type sets and unfold it as a neural network module to enable end-to-end training. 
    Experiments on UFET show that the Neural-PCRF consistently outperforms its backbones with little cost and results in a competitive performance against cross-encoder based SOTA while being \emph{thousands of times} faster. We also find Neural-PCRF effective on a widely used fine-grained entity typing dataset with a smaller type set. We pack Neural-PCRF as a network module that can be plugged onto multi-label type classifiers with ease and release it in \code.
\end{abstract}

 \section{Introduction}
Entity typing assigns semantic types to entities mentioned in the text. The extracted type information has a wide range of applications. It acts as a primitive for information extraction \cite{2phaseNER} and spoken language understanding \cite{coucke2018snips}, and assists in more complicated tasks such as machine reading comprehension \cite{joshi2017triviaqa} and semantic parsing \cite{yavuz-etal-2016-improving}.
During its long history of development, the granularity of the type set for entity typing and recognition changes from coarse (sized less than 20) \cite{conll03,06ontonotes}, to fine (sized around 100) \cite{bbn,figer,ontonotes,ding-etal-2021-nerd}, to ultra-fine and free-form (sized 10k) \cite{ufet}. The expansion of the type set reveals the diversity of real-world entity categories and the importance of a finer-grained understanding of entities in applications \cite{ufet}.

The increasing number of types results in difficulties in predicting correct types, so a better understanding of the entity types and their correlations is needed. Most previous works solve an $N$-type entity typing problem as $N$ independent binary classifications \cite{figer,ontonotes,ufet,onoe-durrett-2019-learning,ding2021prompt,dfet,lite}. However, types are highly correlated and hence they should be predicted in a joint manner. As an example, when `{\em Joe Biden}' is inferred as a {\it president}, it should also be inferred as a {\it politician}, but not a {\it science fiction}. Type correlation is partially specified in fine-grained entity typing (FET) datasets by a two or three-level type hierarchy, and is commonly utilized by a hierarchy-aware objective function \cite{ren2016label, jin-etal-2019-fine, xu-barbosa-2018-neural}. However, type hierarchies cannot encode type relationships beyond strict containment, such as similarity and mutual exclusion \cite{onoe-etal-2021-modeling}, and could be noisily defined \cite{wu2019modeling} or even unavailable (in ultra-fine entity typing (UFET)). Many recent works handle these problems by embedding types and mentions into special spaces such as the hyperbolic space \cite{lopez-strube-2020-fully} or box space \cite{onoe-etal-2021-modeling} that can be trained to latently encode type correlations without a hierarchy. Although these methods are expressive for modeling type correlations, they are constrained by these special spaces and thus incapable of being combined with modern entity typing backbones such as prompt learning \cite{ding2021prompt} and cross-encoder \cite{lite}, and cannot integrate prior type semantics.

In this paper, we present an efficient method that expressively models type correlations while being backbone-agnostic. We formulate the UFET and FET problems under a classical undirected graphical model (UGM) \cite{pgm} called pairwise conditional random field (PCRF) \cite{pcrf2005}. In PCRF, types are binary variables that are not only unarily influenced by the input but also pairwisely relate to all the other type variables. We formulate the unary potentials using the type logits provided by any modern backbone such as prompt learning. To compose the pairwise potentials sized $O(4N^2)$ ($N$ is the number of types, 10k for UFET), we use matrix decomposition, which is efficient and able to utilize prior type semantic knowledge from word embeddings \cite{glove}. Exact inference on such a large and dense UGM is intractable, and therefore we apply mean-field variational inference (MFVI) for approximate inference. Inspired by \citet{crfrnn}, we unfold the MFVI as a recurrent neural network module and connect it with the unary potential backbone to enable end-to-end training and inference. We call our method Neural-PCRF (NPCRF). Experiments on UFET show that NPCRF consistently outperforms the backbones with negligible additional cost, and results in a strong performance against cross-encoder based SOTA models while being \emph{thousands of times} faster. We also found NPCRF effective on a widely used fine-grained entity typing dataset with a smaller type set. We pack NPCRF as a network module that can be plugged onto multi-label type classifiers with ease and release it in \code.

\begin{figure*}[t]
    \centering
    \includegraphics[width=0.98\textwidth]{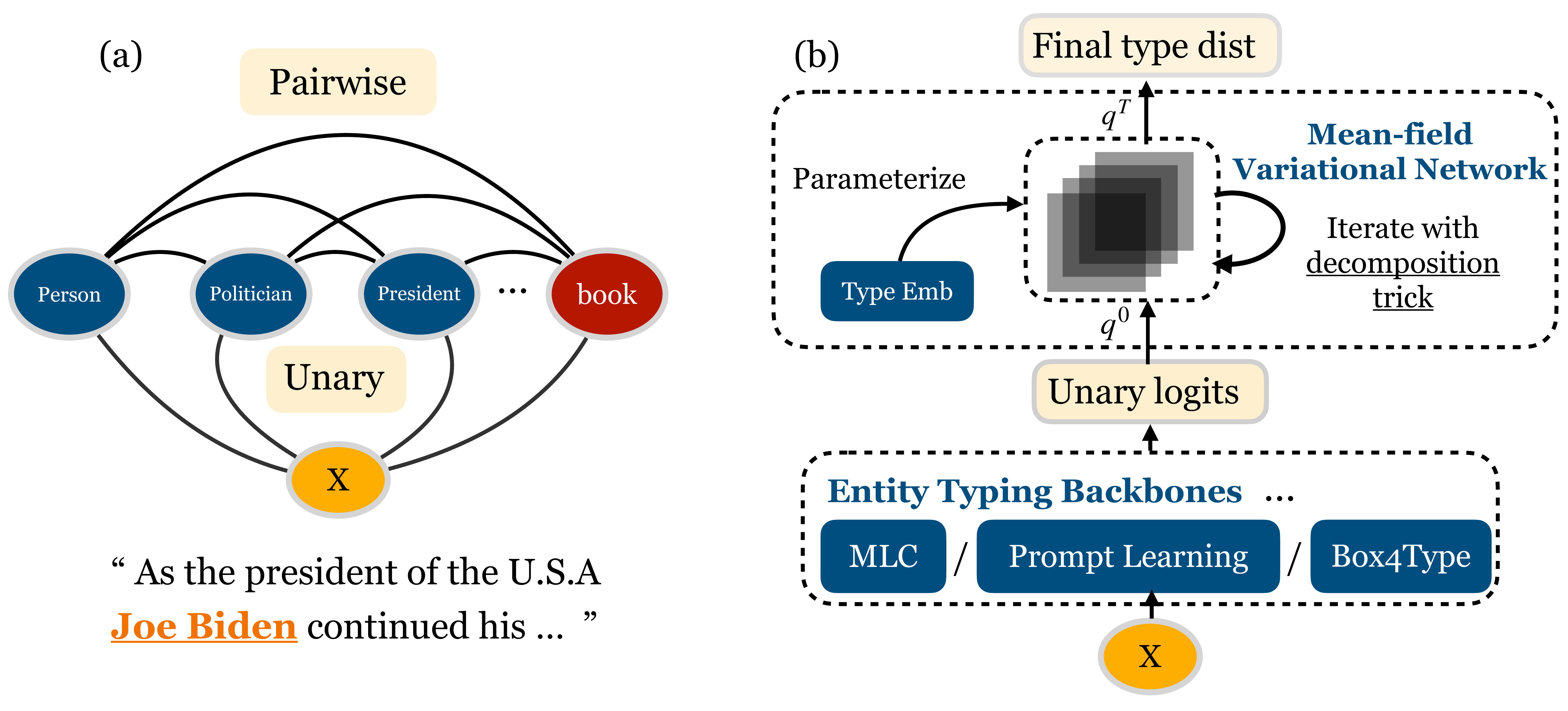}
    \caption{(a) Our pairwise CRF (PCRF) of ultra-fine entity typing. $X$ is the entity mention and its context. (b) The neural network architecture corresponding to the Neural-PCRF.}
    \label{fig:all}
\end{figure*}
\section{Related Work}
\subsection{Fine-grained and Ultra-Fine Entity Typing}
Fine-grained \cite{ontonotes,figer,bbn} (FET) and ultra-fine \cite{ufet} entity typing (UFET) share common challenges. Researchers put lots of effort into better utilizing the distantly and manually labeled training data, including rule-based denoising \cite{ontonotes}, multi-round automatic denoising \cite{onoe-durrett-2019-learning,dfet,mlmet}, and partial label embedding \cite{ren2016label}. Some recent works introduce better entity typing backbones compared with commonly used multi-label classifiers \cite{ontonotes,ufet,onoe-durrett-2019-learning}, such as prompt learning \cite{ding2021prompt}, cross-encoders with indirect supervision from natural language inference \cite{lite} and using a graph neural network to enhance mention representations by sibling mentions for FET \cite{chen-etal-2022-learning-sibling}. Another line of work concentrates on modeling the given or latent type correlations in entity typing, which is more relevant to our work. \citet{ren2016label, jin-etal-2019-fine, xu-barbosa-2018-neural} leverage a given type hierarchy through a hierarchy-aware loss for FET. However, in UFET, the type hierarchy is not given. Therefore various special type embeddings are designed to learn the latent type correlation during training.
\citet{xiong-etal-2019-imposing} utilize graph embedding of a label co-occurrence graph counted from training data, \citet{lopez-strube-2020-fully} embeds entity mentions, contexts and labels into the hyperbolic space that can latently encodes tree structures, and \citet{onoe-etal-2021-modeling} embeds those into the box space that encodes label dependency through the topology of hyperrectangles. Unlike them, we treat types as random variables under an undirected probabilistic graphical model, and only require the type logits, therefore our neural PCRF is backbone-agnostic. \citet{liu-etal-2021-fine} propose a label reasoning network (LRN) that generates types in an auto-regressive manner, while our method decodes types in parallel and can be combined with prompt-based backbones which LRN cannot do.

\subsection{Pairwise Conditional Random Field}
Pairwise conditional random field \cite{pcrf2005} is a classical undirected graphical model proposed for modeling label correlations. In the deep learning era, PCRF was first found effective when combined with a convolutional neural network (CNN) \cite{cnn} for semantic segmentation in computer vision \cite{crfrnn,pcrf1,pcrf2,pcrf3,fw2021}, in which it was used to encourage adjacent pixels being segmented together. In contrast to its popularity in computer vision, PCRF is much less explored in natural language processing.
\citet{pcrf2005} and \citet{pcrfT} apply PCRF on an n-gram feature based non-neural sentence classifier with up to 203 classes. Besides the difference in tasks, numbers of classes and backbones, our method is different from theirs in two main aspects: (1) We unfold mean-field approximate inference of PCRF as a recurrent neural network module for efficient end-to-end training and inference, while they use the `supported inference method' \cite{pcrf2005} which is intractable for large type sets and incompatible with neural backbones (2) We design the parameterization of the pairwise potential function based on the technique of matrix decomposition to accelerate training and inference, and encode type semantics which is important for entity typing \cite{lite} and sentence classification \cite{mueller2022label}. Our parameterization also conforms to intrinsic properties of pairwise potentials (explained in Sec. \ref{sec:pairwise}). \citet{hu-etal-2020-investigation} investigates different potential function variants for linear-chain CRFs for sequence labeling, while we design pairwise potential functions for PCRF.
\section{Method}
\subsection{Problem Definition} 
Entity typing datasets consist of $M$ entity mentions $m_i$ with their corresponding context sentences $c_i$: $\mathcal{D}=\{(m_1,c_1),\cdots, (m_M, c_M)\}$, and a type set $\mathcal{Y}$ of size $N$. 
The task of entity typing is to predict the types $\bm{y}_i^p$ of the entity mention $m_i$ in the given context $c_i$, where $\bm{y}_i^p$ is a subset of the type set. The number of gold types $|\bm{y}_i^g|$ could be larger than one in most entity typing datasets. The average number of gold types per instance $\text{avg}({\vert \bm{y}_i^g \vert})$ and the size of the type set $|\mathcal{Y}|$ vary in different datasets.

\subsection{PCRF for Entity Typing}
We first describe our pairwise conditional random field for the entity typing \cite{pcrf2005}, as shown in Fig. \ref{fig:all}(a). $x$ denotes a data point $(m, c) \in \mathcal{D}$. $Y_j \in \{0, 1\}$ denotes the binary random variable for the $j$-th type in type set $\mathcal{Y}$. The type variables $Y_{1:N}$ are unarily connected to the input, model how likely a type is given the input, and pairwisely connect a type variable with all other type variables to model type correlations.
Let $\bm{y} \in Y_1 \times Y_2 \times \cdots \times Y_N$ be an assignment of all the type variables. The probability of $\bm{y}$ given a data point $x_i$ under a conditional random field can be factorized as:
\[
    p(\bm{y} \vert x)=\frac{1}{Z} \exp{\Big( \sum_{j} \theta_u(y_j ; x) + \sum_{j < k} \theta_p(y_j, y_k) \Big)}
\]
where $\theta_u, \theta_p$ are scoring functions for unary and pairwise edges, and $Z$ is the partition function for normalization.

\subsection{Unary Potential Function}
\label{sec:backbone}
As illustrated in Figure \ref{fig:all}(b), we parameterize the log unary potential function $\theta_u(y_j;x)$ using modern entity typing backbones based on pretrained language models (PLM)\footnote{We use PLMs from \url{https://huggingface.co/}}.We introduce two of the backbones below. 
\paragraph{Multi-label Classifier (MLC)} Given a data point $(m, c)$, we formulate the input as:
\[
[\text{CLS}] \  c \  [\text{SEP}] \   m \ [\text{SEP}] 
\]
and feed it into RoBERTa-large \cite{liu2019roberta} to obtain the $[\text{CLS}]$ embedding $\bm{h}$. Then we obtain the unary type scores as follows.
\[
\begin{aligned}
\theta_u(y_j=1;x) &= \text{Linear}(\bm{h})[j] \\
\theta_u(y_j=0;x) &= 0
\end{aligned}
\]
\paragraph{Prompt Learning (PL)} We adopt prompt learning for entity typing \cite{ding2021prompt,dfet} as our second backbone. We feed:
\[
[\text{CLS}] \  c \  [\text{P}_1] \   m \ [\text{P}_2]  \ [\text{P}_3]  \ [\text{MASK}]  \ [\text{SEP}] 
\]
into Bert-cased \cite{devlin2018bert}, where $[\text{P}_1],[\text{P}_2],[\text{P}_3]$ are trainable soft prompt tokens. We use a verbalizer $\mathcal{V}$ to map types to subwords, $\mathcal{V}_y$ denotes the subwords corresponding to type $y \in \mathcal{Y}$, e.g., $\mathcal{V}_\text{\emph{living\_thing}}=\{\text{\emph{`living', `thing'}}\}$. We obtain the unary score of choosing type $y_j$ using the averaged masked language model logits of $\mathcal{V}_{y_j}$:
\[
\theta_u(y_j=1;x) = \sum_{w \in \mathcal{V}_{y_j}} s([\text{MASK}]=w) / \vert \mathcal{V}_{y_j} \vert
\]
Similarly, we set $\theta(y_j=0;x) = 0$.

\subsection{Pairwise Potential Function}
\label{sec:pairwise}
\begin{table}[t]
\renewcommand\arraystretch{1.15}
\centering
\scalebox{0.75}{
\begin{tabular}{ccc} \toprule
   potential               &       encode          &    property \\ \midrule
$\Theta_{11}(j,k) = \theta_p(y_j=1,y_k=1)$& co-occurrence & $\Theta_{11}=\Theta_{11}^T$ \\
$\Theta_{00}(j,k) =\theta_p(y_j=0,y_k=0)$ & co-absence   & $\Theta_{00}=\Theta_{00}^T$\\
$\Theta_{10}(j,k) =\theta_p(y_j=1,y_k=0)$ & co-exclusion & $\Theta_{10}=\Theta_{01}^T$ \\
$\Theta_{01}(j,k) =\theta_p(y_j=0,y_k=1)$ & co-exclusion & $\Theta_{01}=\Theta_{10}^T$ \\ \bottomrule
\end{tabular}}
\caption{Properties of the four log potential matrices.}
\label{tab:pairwise}
\end{table}

Pairwise log potential $\theta_p(y_j, y_k)$ encoding type correlations should naturally satisfy the properties shown in Table~\ref{tab:pairwise}.
Specifically, $\Theta_{11}$ and $\Theta_{00}$ are symmetric matrices and encode co-occurrence and co-absence respectively, while $\Theta_{10}=\Theta_{01}^T$ and are asymmetric ( e.g., ``not a person but a president'' $\not=$ ``not a president but a person''). Directly parameterizing these 4 potential matrices \cite{pcrf2005,pcrfT} ignores these intrinsic properties and results in an unbearable number of model parameters for datasets with a large type set (e.g., 400M parameters for 10331-type UFET, which is more than Bert-large).

To tackle these problems, we parameterize the pairwise potential based on matrix rank decomposition, i.e., we represent each $N \times N$ matrix as the product of two $N \times R$ matrices, where $R$ is the number of ranks.
Crucially, we use pretrained word embedding to derive the two $N \times R$ matrices, so that they can encode type semantics.
Specifically, we obtain a type embedding matrix $\bm{E} \in \mathbb{R}^{N \times 300}$ based on 300-dimension GloVe embedding \cite{glove}. For a type phrase consisting of multiple words (e.g., \emph{``living\_thing''}), we take the average of the word embeddings. We then transform $\bm{E}$ into two embedding spaces encoding \emph{``occurrence''} and \emph{``absence''} respectively using two feed-forward networks.
\begin{equation}
\begin{aligned}
\text{FFN}(\bm{E}) &=  \bm{W_2} (\tanh( \text{dropout}(\bm{W_1} \bm{E})) \\
\bm{E}_0 &= \text{FFN}_1(\bm{E}) \in \mathbb{R}^{N \times R} \\
\bm{E}_1 &= \text{FFN}_2(\bm{E}) \in \mathbb{R}^{N \times R} \\
\end{aligned}
\label{eq:1}
\end{equation}

To enforce the intrinsic properties of the four matrices, we parameterize them as follows:
\begin{equation}
\begin{aligned}
\Theta_{00} &=  \bm{E}_0 \bm{E}_0^T &
\Theta_{11} &= \bm{E}_1 \bm{E}_1^T\\
\Theta_{01} &= - \bm{E}_0 \bm{E}_1^T &
\Theta_{10} &= - \bm{E}_1 \bm{E}_0^T
\end{aligned}
\label{eq:2}
\end{equation}
The negative signs in defining $\Theta_{01}, \Theta_{10}$ ensure that they encode co-exclusion, not similarity. As we will show in the next subsection, we do not need to actually recover these large matrices during inference, leading to lower computational complexity.

\subsection{Mean-field Variational Inference}
\label{sec:mfvi}
We aim to infer the best type set given the input.
\[\bm{y}^p=\argmax_{\bm{y}\in Y_1 \times \cdots \times Y_N } p(\bm{y}\vert x)\]
The MAP (maximum-a-posteriori) inference over this large and dense graph is NP-hard. We instead use mean-field variational inference (MFVI) to approximately infer the best assignment $\bm{y}^p$. MFVI approximates the true posterior distribution $p(y_j \vert x)$ by a variational distribution $q(y_j)$ that can be factorized into independent marginals \cite{wainwright2008graphical}. It iteratively minimizes the KL-divergence between $p$ and $q$. We initialize the $q$ distribution by unary scores produced by backbones:
\[
 q^{0}(y_j) = \text{softmax}(\theta_u(y_j;x))
\]
 and derive the MFVI iteration as follows:
\[
\scalemath{1.0}{
\begin{aligned}
& q^{t+1}(y_j=1) \propto \exp \Big(\theta_u(y_j=1;x) +  \\
& \sum_{k \not= j }(\Theta_{11}^{(jk)} \cdot q^t(y_k=1)) + \Theta_{10}^{(jk)} \cdot q^t(y_k=0) \Big) \\
& q^{t+1}(y_j=0) \propto \exp \Big(\theta_u(y_j=0;x) +  \\
& \sum_{k \not= j }(\Theta_{00}^{(jk)} \cdot q^t(y_k=0)) + \Theta_{01}^{(jk)} \cdot q^t(y_k=1) \Big) \\
\end{aligned}}
\]
We rewrite the formulas in the vector form with our parametrization. $\bm{\theta}_{1}^{u} \in \mathbb{R}^N$ denotes the unary score vector of $\theta_u(y_j=1; x)$ for all $j$, and we define $\bm{\theta}_{0}^{u}$ similarly.
$\bm{q}_0^{t}, \bm{q}_1^{t}$ are vectors of the $q$ distributions at the $t$-th iteration.
\[
\begin{aligned}
 & (\bm{q}_0^{0}[j], \bm{q}_1^{0}[j]) = \text{softmax}(\bm{\theta}_0^{u}[j], \bm{\theta}_1^{u}[j]) \\
 & \bm{q}_{1}^{t+1} \propto \exp \Big( \bm{\theta}_1^u  + \bm{E}_1 \big( \bm{E}_1^T \bm{q}^t_1 \big) - \bm{E}_1 \big( \bm{E}_0^T \bm{q}^t_0 \big) \Big) \\
 & \bm{q}_{0}^{t+1} \propto \exp \Big( \bm{\theta}_0^u  + \bm{E}_0 \big( \bm{E}_0^T \bm{q}^t_0 \big) - \bm{E}_0 \big( \bm{E}_1^T \bm{q}^t_1 \big) \Big)
 \end{aligned}
\]

Note that by following these update formulas, we do not need to recover the $N \times N$ matrices, and hence the time complexity of each iteration is $NR$. Since the iteration number $\mathcal{T}$ is typically a small constant ($\mathcal{T} < 7$), the computational complexity of the whole MFVI is $O(NR)$. 
The predicted type set of $x_i$ is obtained by:
\[
\bm{y}^p = \{y_j \ \vert \ \bm{q}_1^\mathcal{T}[j] > 0.5, y_j \in \mathcal{Y} \}
\]
We follow the treatment of MFVI as entropy-regularized Frank-Wolfe \cite{le2021regularized} and introduce another hyper-parameter $\lambda$ to control the step size of each update. Let $\bm{q}^t =[\bm{q}^{t}_0; \bm{q}^{t}_1]$,
\[
\bm{q}^{t+1} = \bm{q}^{t} + \lambda (\bm{q}^{t+1} - \bm{q}^{t}  )
\]
\subsection{Unfolding Mean-field Variational Inference as a Recurrent Neural Network}
We follow \citet{crfrnn} and treat the mean-field variational inference procedure with a fixed number of iterations
 as a recurrent neural network (RNN) parameterized by the pairwise potential functions. As shown in the top part of Figure \ref{fig:all}(b), the initial hidden state of the RNN is the type distribution $\bm{q}^0$ produced by the unary logits, and the final hidden states $\bm{q}^\mathcal{T}$ after $\mathcal{T}$ iterations are used for end-to-end training and inference.
 
 \subsection{Training Objective and Optimization} 
We use the Binary Cross Entropy loss for training:
\[
\scalemath{1.0}{
\begin{aligned}
\mathcal{L} = -\frac{1}{N} \sum_{j=1}^N \Big( \alpha \cdot y_{j}^g \log \bm{q}^{\mathcal{T}}_1[j] + \\
(1-y_{j}^g) \log (1-\bm{q}^{\mathcal{T}}_1[j]) \Big)
\end{aligned}
}
\]
$\mathcal{L}$ is the loss of each instance $x$, $y_{j}^g \in \{0, 1\}$ is the gold annotation of $x$ for type $y_j$. We follow previous works \cite{ufet,mlmet} and use $\alpha$ as a weight for the loss of positive types. 
We train the pretrained language model, label embedding matrix $\bm{E}$ and FFNs.
We use AdamW \cite{loshchilov2018fixing} for optimization.
\section{Experiment}
\subsection{Datasets and Experimental Settings}
We mainly evaluate our NPCRF method on the ultra-fine entity typing dataset: UFET \cite{ufet}. We also conduct experiments on the augmented version \cite{ufet} of OntoNotes \cite{ontonotes} to examine if our method also works for datasets with smaller type sets and fewer gold types per instance. We show the dataset statistics in Table \ref{tab:stat}. Note that UFET also provides 25M distantly labeled training data extracted by linking to KB and parsing. We follow recent works \cite{dfet,liu-etal-2021-fine} and only use the manually annotated 2k data for training. We use standard metrics for evaluation: for UFET, we report macro-averaged precision (P), recall (R), and F1; for OntoNotes, we report macro-averaged and micro-averaged F1. We run experiments three times and report the average results.

\begin{table}[h]
\scalebox{0.8}{
\centering
\begin{tabular}{ccccc} 
\toprule
dataset & $\vert \mathcal{Y} \vert$      & $\text{avg}(\vert \bm{y}_i^g \vert)$ & train/dev/test & hierarchy \\ \midrule
UFET    & 10331       & 5.4   & 2k/2k/2k       & No        \\
OntoNotes  & 89          & 1.5   &   0.8M/2k/9k             & 3-layer       \\ \bottomrule
\end{tabular}}
\caption{$\vert \mathcal{Y} \vert$ denotes the size of the type set, $\text{avg}(\vert \bm{y}_i^g \vert)$ denotes the average number of gold types per instance.}
\label{tab:stat}
\end{table}

\subsection{Baseline Methods}
\begin{table}[t]
\centering
\scalebox{0.8}{
\begin{tabular}{llll} \toprule
\bf \textsc{Model}                & \bf \textsc{P}    & \bf \textsc{R}    &  \bf \textsc{F1} \\ \midrule
\multicolumn{4}{l}{\emph{non-PLM model for reference}}        \\
{\bf \textsc{UFET-BiLSTM}}  \cite{ufet} $^\dagger$    & 48.1 & 23.2 & 31.3 \\
{\bf \textsc{LabelGCN}} \cite{xiong-etal-2019-imposing} $^{\dagger \Diamond}$       & 50.3 & 29.2 & 36.9 \\ \hline \midrule
\multicolumn{4}{l}{\emph{PLM-based models}}        \\
{\bf \textsc{LDET}} \cite{onoe-durrett-2019-learning}  $^\dagger$        & 51.5 & 33.0 & 40.1 \\
{\bf \textsc{Box4Types}} \cite{onoe-etal-2021-modeling} $^{\dagger \Diamond}$ & 52.8 & 38.8 & 44.8 \\
{\bf \textsc{LRN} }  {\cite{liu-etal-2021-fine}} $^\Diamond$               & 54.5 & 38.9 & 45.4 \\
{\bf \textsc{MLMET}}  {\cite{mlmet}} $^\dagger$   & 53.6 & 45.3 & 49.1 \\
{\bf \textsc{LITE+L}}  \cite{lite}             & 48.7 & 45.8 & 47.2 \\
{\bf \textsc{DFET}}    \cite{dfet}             & 55.6 & 44.7 & 49.5 \\
{\bf \textsc{LITE+NLI+L}} \cite{lite} $^\dagger$ & 52.4 & 48.9 & {\bf 50.6} \\ \hline \midrule
\multicolumn{4}{l}{\emph{PLM-based models w/ or w/o NPCRF}}        \\
\bf \textsc{MLC-RoBerta}                  & 47.8 & 40.4 & 43.8 \\
\bf \textsc{MLC-RoBerta w/ NPCRF}             & 48.7 & 45.5 & 47.0  \\ [0.5ex] \hdashline \\[-2ex]
\bf \textsc{LabelGCN-RoBerta}$^\Diamond$     & 51.2 & 41.0 & 45.5 \\
\bf \textsc{LabelGCN-RoBerta w/ NPCRF}    & 48.7 & 45.9 & 47.3  \\ [0.5ex] \hdashline \\[-2ex]
{\bf \textsc{PL-Bert-base}}  \cite{ding2021prompt}       & 57.8 & 40.7 & 47.7 \\
\bf \textsc{PL-Bert-base w/ NPCRF}    & 52.1 & 47.5 & 49.7  \\ [0.5ex] \hdashline \\[-2ex]
\bf \textsc{PL-Bert-large}        & 59.3 & 42.6 & 49.6 \\
\bf \textsc{PL-Bert-large w/ NPCRF}   & 55.3 & 46.7 & {\bf 50.6} \\ \bottomrule
\end{tabular}}
\caption{Macro-averaged UFET result. {\bf \textsc{LITE+L}} is LITE without NLI pretraining, {\bf \textsc{LITE+L+NLI}} is the full LITE model,  {\bf \textsc{LabelGCN-RoBerta}} denotes our implementation of LabelGCN with RoBerta-large as mention encoder. 
Methods marked by $\dagger$ use either distantly labeled training data or additional pretraining tasks and $\Diamond$ marker denotes method focusing on modeling label correlations.}
\label{tab:ufet}
\end{table}

{\bf \textsc{MLC-RoBerta}} and {\bf \textsc{PL-Bert}} introduced in Sec. \ref{sec:backbone} are natural baselines. We compare their performances with and without NPCRF. \\
{\bf \textsc{UFET} }\cite{ufet} \ A multi-label linear classifier (MLC) with a backbone using BiLSTM, GloVe, and CharCNN to encode the mention. \\
{\bf \textsc{LDET} }\cite{onoe-durrett-2019-learning} \ An MLC with Bert-base-uncased and ELMo \cite{elmo} and trained by 727k examples automatically denoised from the distantly labeled UFET. \\
{\bf \textsc{LabelGCN} }\cite{xiong-etal-2019-imposing} \ An MLC with BiLSTM and multi-head self-attention to encode the mention and context, and a GCN running on a fixed co-occurrence type graph to obtain better type embedding. Type scores are dot-product of mention and type embedddings. For fair comparison, we replace their mention encoder with RoBerta-large. To our knowledge, LabelGCN cannot be directly combined with prompt learning. \\
{\bf \textsc{Box4Type} }\cite{onoe-etal-2021-modeling} use Bert-large as backbone and project mentions and types to the box space for training and inference; trained on the same 727k data \cite{onoe-durrett-2019-learning}. \\
{\bf \textsc{LRN} }\cite{liu-etal-2021-fine} generate types using Bert-base and an LSTM decoder in a seq2seq manner, and use 2k manually labeled data for training.   \\
{\bf \textsc{MLMET} }\cite{mlmet} \ A multi-label linear classifier using Bert-base, first pretrained by the distantly-labeled data augmented by masked word prediction, then finetuned and self-trained on the 2k human-annotated data. \\
{\bf \textsc{DFET} }\cite{dfet} \ A 3-round automatic denoising method for 2k mannually labeled data, using {\bf \textsc{PL-Bert}} as backbone. \\
{\bf \textsc{LITE} }\cite{lite} \ Previous SOTA system that formulates entity typing as natural language inference, treating $x_i$ as premise and types as hypothesis, concatenating them and feeding them into RoBerta-large to score types. LITE models the correlation better between the input and types, and has great zero-shot performance but needs to concatenate $x_i$ with all the $N$ types, resulting in very slow inference. LITE is pretrained on MNLI \cite{mnli} and trained on 2k annotated data, Its authors find that the performance drops when using distantly labeled data.

\subsection{UFET Result}
As shown in Table \ref{tab:ufet}, our {\bf \textsc{NPCRF}} can be integrated with various entity typing backbones and enhance their performance to SOTA on UFET.

{\bf \textsc{MLC-RoBerta w/ NPCRF}} improves the basic {\bf \textsc{MLC-RoBerta}} backbone by $+3.2$ F1 score and reaches the best performance among models using the MLC architecture except for {\bf \textsc{MLMET}} which uses millions of distantly labeled data. {\bf \textsc{LabelGCN}} \cite{xiong-etal-2019-imposing} utilizing fixed type co-occurrence information to obtain type embedding is still effective with the replaced {\bf \textsc{RoBerta}} encoder and improves F1 by $+1.7$ over {\bf \textsc{MLC-RoBerta}} while our method produce a further $+1.8$ F1 improvement, because {\bf \textsc{NPCRF}} models not only co-occurrence (in $\Theta_{11}$), but also semantic co-absence and co-exclusion in our pairwise potentials, and it also explicitly updates these potentials during training rather than using fixed type co-occurrence information.

Prompt-learning based backbones \cite{ding2021prompt,dfet} such as {\bf \textsc{PL-BERT}} are already strong in UFET, and our method 
further improves {\bf \textsc{PL-BERT-Base}} by $+1.9$ F1 and reaches $49.7$ F1 which is slightly better than {\bf \textsc{MLMET}} and {\bf \textsc{DFET}} and is the SOTA performance for Bert-base models, on par with performance of Bert-large models. Also worth noting is that our method is trained in a simpler single-round end-to-end manner compared with other competitors requiring multi-round training \cite{mlmet} and denoising \cite{dfet} procedure. 
For prompt-learning powered by Bert-large models, {\bf \textsc{NPCRF}} boosts the performance of {\bf \textsc{PL-Bert-large}} by $1.0$ F1, and results in performance on par with the previous SOTA system {\bf \textsc{LITE+NLI+L}} with much faster inference speed (discussed in Sec. \ref{sec:speed}). Note that the improvement is smaller on large models because large models are stronger in inferring types without pairwise potentials from limited and noisy contexts. We also observe that models with {\bf \textsc{NPCRF}} tend to predict more types and therefore have higher recalls and sometimes lower precisions. It is possibly because {\bf \textsc{NPCRF}} can use type dependencies to infer additional types that are not directly supported by the input. We will discuss it in detail in Sec. \ref{sec:case}.

\subsection{FET Result}
We present the performance of {\bf \textsc{NPCRF}} on the augmented OntoNotes dataset in Table \ref{tab:onto}. The results show that {\bf \textsc{MLC-RoBerta w/ NPCRF}} outperforms {\bf \textsc{MLC-RoBerta}} by $+1.5$ in macro-F1 and $+1.7$ in micro-F1, and reaches competitive performances against a recent SOTA system {\bf \textsc{DFET}} focusing on denoising data. In general, we find our method still effective, but the improvement is less significant compared with UFET, especially for {\bf \textsc{PL-Bert}}. One possible reason is that the average number of types per instance in OntoNotes is 1.5, and therefore the type-type relationship is less important. Another possible reason is that some type semantics are already covered in the prompt-based method through the verbalizer, so our method fails to boost the performance of {\bf \textsc{PL-Bert}}. 

\begin{table}[t]
\centering
\scalebox{0.75}{
\begin{tabular}{llcc} \toprule
\bf \textsc{Model}                & \bf \textsc{Ma-F1}    &  \bf \textsc{Mi-F1} \\ \midrule
\multicolumn{4}{l}{\emph{PLM-based models}}        \\
{\bf \textsc{LDET}} \cite{onoe-durrett-2019-learning}  $^\dagger$         & 84.5 & 79.2  \\
{\bf \textsc{Box4Types}} \cite{onoe-etal-2021-modeling} $^{\dagger \Diamond}$  & 77.3 & 70.9 \\
{\bf \textsc{LRN} }  {\cite{liu-etal-2021-fine}} $^\Diamond$                  & 84.8 & 80.1 \\
{\bf \textsc{MLMET}}  {\cite{mlmet}} $^\dagger$     & 85.4 & 80.4  \\
{\bf \textsc{LITE}}  \cite{lite}                & 86.6 & 81.4  \\
{\bf \textsc{DFET}}    \cite{dfet}                 & \textbf{87.1} & 81.5  \\ \hline \midrule
\multicolumn{4}{l}{\emph{PLM-based models w/ or w/o NPCRF}}        \\
\bf \textsc{MLC-RoBerta}                  & 85.4 & 80.2  \\
\bf \textsc{MLC-RoBerta w/ NPCRF}         & 86.9 & \textbf{81.9}  \\ [0.5ex] \hdashline \\[-2ex]
\bf \textsc{LabelGCN-RoBerta} $^\Diamond$      & 85.5 & 80.2 &  \\
\bf \textsc{LabelGCN-RoBerta w/ NPCRF}    & 86.2 & 81.3 &   \\ [0.5ex] \hdashline \\[-2ex]
{\bf \textsc{PL-Bert-Base}} \cite{ding2021prompt}                 & 84.8 &  79.6\\
\bf \textsc{PL-Bert-Base w/ NPCRF}         & 85.2 & 80.0  \\ \bottomrule
\end{tabular}}
\caption{Results on the augmented OntoNotes datasets. {\bf \textsc{Ma}} and {\bf \textsc{Mi}} are abbreviations of macro and micro. Methods marked by $\dagger$ use either distantly labeled training data or additional pretraining tasks and the $\Diamond$ marker denotes methods modeling label correlations. }
\label{tab:onto}
\end{table}

\section{Analysis}
\subsection{Ablation Study}
We show the ablation study in Table \ref{tab:ablation}. It can be seen that: (1) Performance drops when we randomly initialize label embedding $\bm{E}$ (denoted by {\bf {- w/o GloVe}}), which indicates that GloVe embedding contains useful type semantics and helps build pairwise type correlations. (2) Performance drops when we remove the hidden layer and the $\tanh$ nonlinearity in {\bf FFN}s (Eq. \ref{eq:1}), showing the benefit of expressive parameterization. (3) When we remove the entire {\bf FFN} layers and use $\bm{E}$ to parameterize the four matrices in (Eq. \ref{eq:2}) model training fails to converge, showing that a reasonable parameterization is important.
\begin{table}[t]
\centering
\scalebox{0.9}{
\begin{tabular}{lcccc} \toprule
\bf \textsc{Model}                & \bf \textsc{Ma-P}    &  \bf \textsc{Ma-R}  &  \bf \textsc{Ma-F1} \\ \midrule
\bf {PL-Bert w/ NPCRF}     & 52.1 & 47.5 & {\bf 49.7} \\
\bf {- w/o GloVe}     & 48.5 & 49.1 & 48.8  \\
\bf {- w/o FFN hidden}      & 51.1 & 47.0 & 49.0  \\ 
\bf {- w/o FFN}      & - & - & -  \\ 
\bf {- w/o NPCRF}     & 57.8 & 40.7 & 47.7  \\ \bottomrule
\end{tabular}}
\caption{Ablation study on UFET test set, ``-'' means training fails to converge.}
\label{tab:ablation}
\end{table}

\subsection{Performance on Coarser Granularity}
We evaluate the performance of {\bf \textsc{NPCRF}} on type sets with coarser granularity. UFET \cite{ufet} splits the types into coarse-grained, fine-grained and ultra-fine-grained types. We discard ultra-fine-grained types to create the fine-grained setting (130 types), and further discard the fine-grained types to create the coarse-grained setting (9 types). As shown in Table \ref{tab:coarser}, {\bf \textsc{NPCRF}} still has positive effect in these two settings, while {\bf \textsc{LabelGCN}} does not.

\begin{table}[t]
\centering
\scalebox{0.8}{
\begin{tabular}{lcccc} \toprule
\bf \textsc{Model}                & \bf \textsc{Mi-P}    &  \bf \textsc{Mi-R}  &  \bf \textsc{Mi-F1} \\ \midrule
\multicolumn{4}{l}{\emph{coarse-grained, 9-class, $\text{avg}(\vert \bm{y}_i^g \vert)=0.95$}}        \\
\bf {\textsc{MLC-RoBERTa}}     & 76.5 & 75.8 & 76.2  \\
\bf {\textsc{LabelGCN-RoBERTa}}      & 77.5 & 74.7 & 76.1  \\ 
\bf {\textsc{MLC-RoBERTa w/ NPCRF}}     & 78.0 & 76.0 &  {\bf 77.0} \\ \midrule
\multicolumn{4}{l}{\emph{fine-grained, 130-class, $\text{avg}(\vert \bm{y}_i^g \vert)=1.61$}}        \\
\bf {\textsc{MLC-RoBERTa}}     & 62.8 & 67.3 & 64.9  \\ 
\bf {\textsc{LabelGCN-RoBERTa}}      & 59.7 & 69.6 & 64.3  \\ 
\bf {\textsc{MLC-RoBERTa w/ NPCRF}}     & 62.7 & 68.8 & {\bf 65.6} \\ \bottomrule
\end{tabular}}
\caption{Performances on coarser granularity.}
\label{tab:coarser}
\end{table}

\begin{figure}[h]
     \centering
     \begin{subfigure}[h]{0.4 \textwidth}
         \centering
         \includegraphics[width=\textwidth]{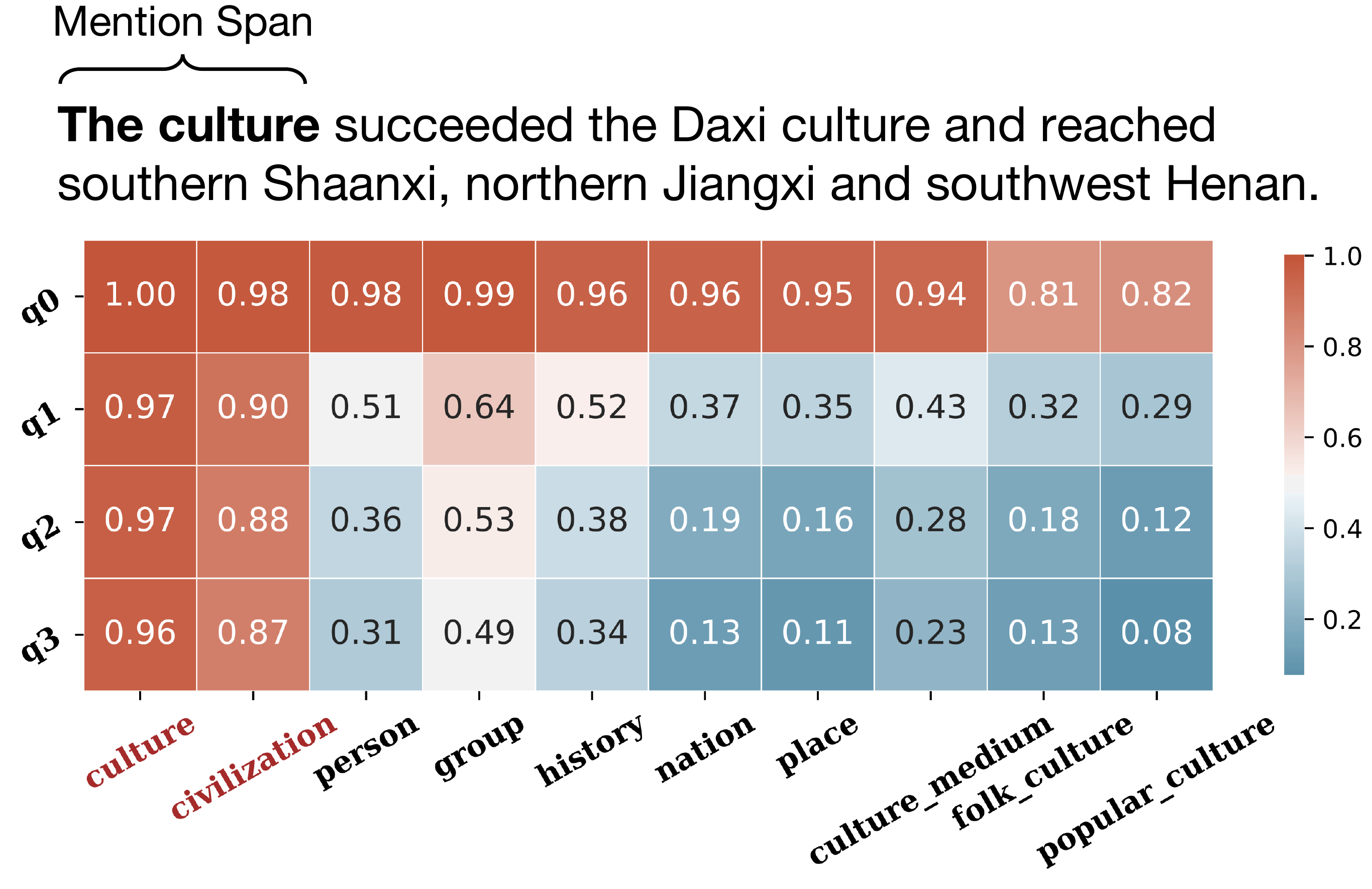}
         \caption{}
         \label{fig:c1}
     \end{subfigure}
     \vfill
     \begin{subfigure}[h]{0.4\textwidth}
         \centering
         \includegraphics[width=\textwidth]{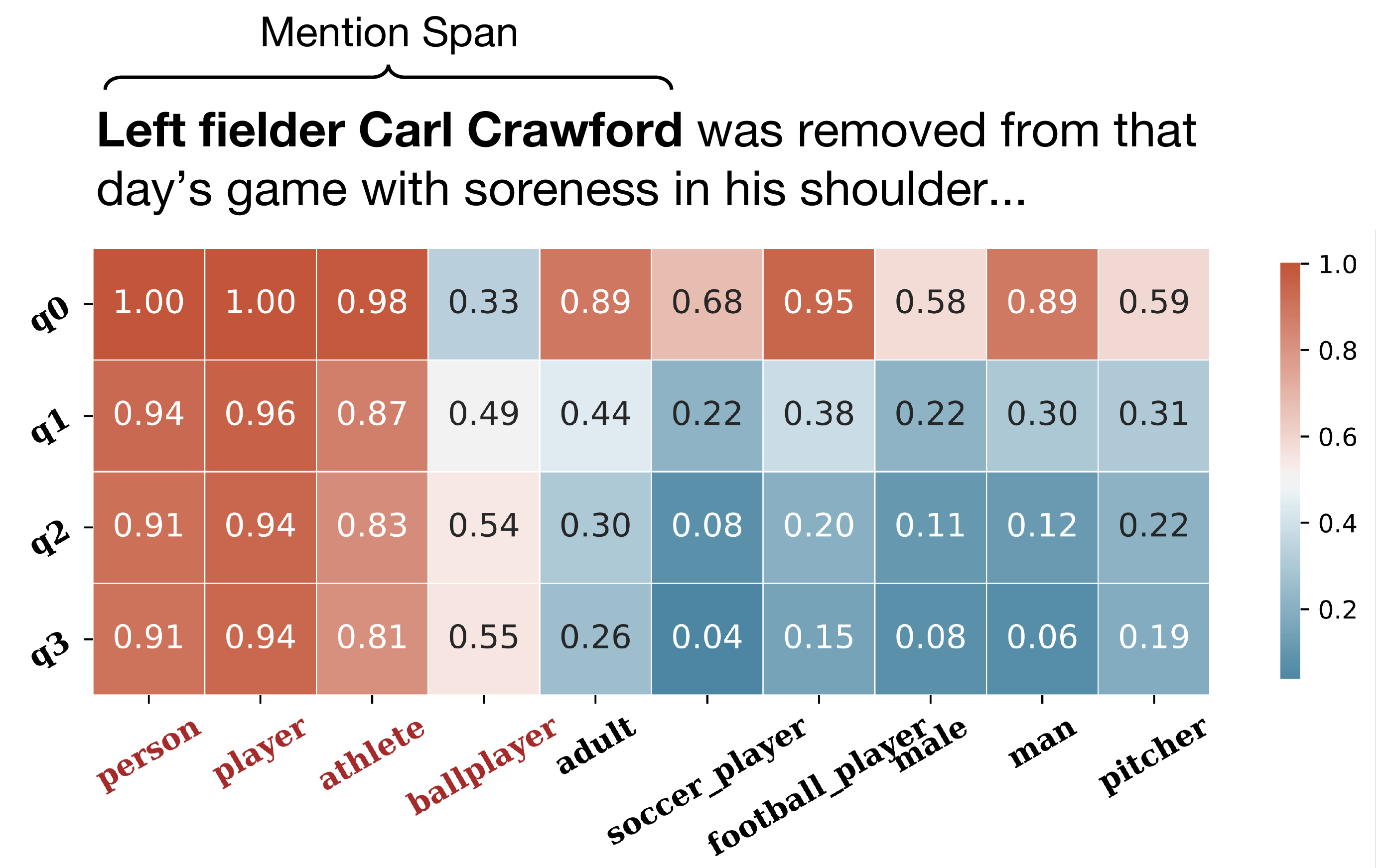}
         \caption{}
         \label{fig:c2}
     \end{subfigure}
     \vfill
     \begin{subfigure}[h]{0.4\textwidth}
         \centering
         \includegraphics[width=\textwidth]{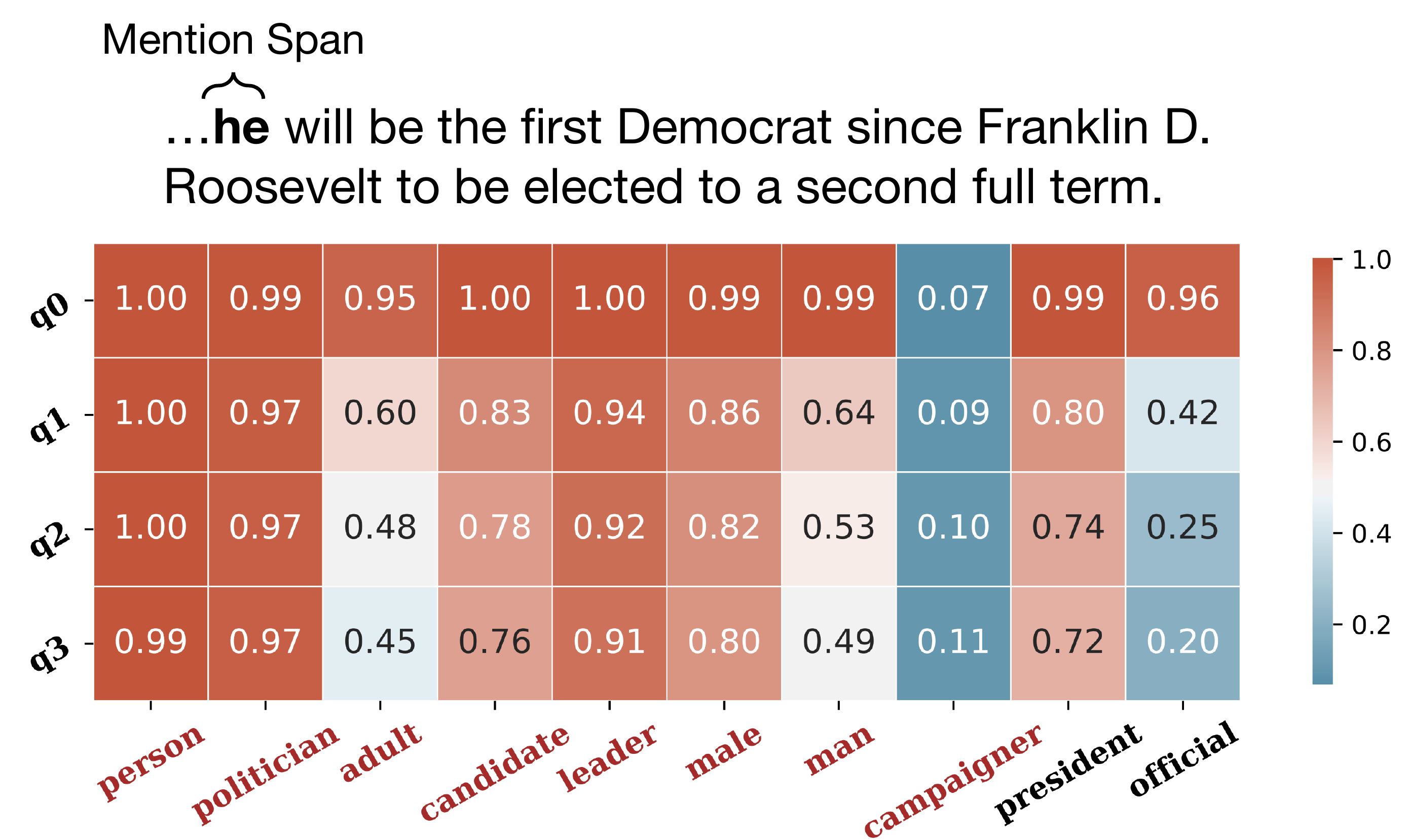}
         \caption{}
         \label{fig:case_3}
     \end{subfigure}
    \caption{Two MFVI cases. We show, from top to bottom, how the type probabilities $\bm{q}_1^t$ change with iteration given the mention and its context. The reddish grids (probability $>0.5$) indicates the chosen types, and gold types are colored red on the x-axis.}
    \label{fig:mfvi}
\end{figure}

\subsection{Model Performance at Each Iteration}
We evaluate the per-iteration performance of {\bf \textsc{PL-Bert w/ NPCRF}} on the test set and show the results in Table \ref{tab:iter}. We obtain the prediction $\bm{y}_{t}^p$ of each iteration $t$ by binarizing $\bm{q}_t$ for each instance: $\bm{y}^p_{t} = \{y_j \ \vert \ \bm{q}_1^{t}[j] > 0.5, y_j \in \mathcal{Y} \}$. The results show that model prediction at iteration 0 (i.e., based solely on unary scores) has a high recall and a very low precision, while
{\bf \textsc{NPCRF}} keeps correcting wrong candidates during consecutive iterations to reach a higher F1 score. We show some concrete cases in Sec. \ref{sec:case}.

\begin{table}[t]
\centering
\scalebox{0.8}{
\begin{tabular}{ccccc} \toprule
\bf \textsc{Iteration}                & \bf \textsc{Ma-P}    &  \bf \textsc{Ma-R}  &  \bf \textsc{Ma-F1} & $\text{avg}(\vert \bm{y}_{i,t}^p \vert)$\\ \midrule
$\bm{q}^0$    & 15.3 & 74.5  & 25.4 & 27.92 \\
$\bm{q}^1$    & 43.2 & 53.6 & 47.8 & 6.52 \\
$\bm{q}^2$    & 50.0 & 48.2 & 49.1 & 4.95 \\ 
$\bm{q}^3$    & 52.1 & 47.4 & 49.7 & 4.65 \\ \bottomrule
\end{tabular}}
\caption{Model performance at each iteration, $\text{avg}(\vert \bm{y}_i^p \vert)$ denotes the average number of predicted types.}
\label{tab:iter}
\end{table}

\subsection{Case Study and Visualization}
\label{sec:case}

\paragraph{MFVI Iterations}
We show the per-iteration predictions of {\bf \textsc{PL-Bert w/ NPCRF}} for two inputs in Figure \ref{fig:mfvi}. As can be seen, NPCRF tends to delete wrong types through iterations, such as \emph{``history'', ``nation''} in Fig. \ref{fig:mfvi}(a) and \emph{``pitcher'', ``soccer\_player''} in Fig. \ref{fig:mfvi}(b). This results in higher precision (as shown in Table \ref{tab:iter}). However, we find NPCRF is also capable of increasing the probabilities of some types, e.g., \emph{``ballplayer''} in the second case. NPCRF may also erroneously deprecate gold types and results in lower recall. As been shown in \ref{fig:mfvi}(c), NPCRF wrongly deletes gold types such as \emph{``adult'', ``man''} while it correctly predicts \emph{``president''} which is not annotated as gold label, and increases the score of \emph{``compaigner''}.

\paragraph{Pairwise Potentials} We show the four learned pairwise potentials in Appendix \ref{app:potential}.

\subsection{Efficiency}
\label{sec:speed}
 We compare the training and inference efficiency of different methods on the UFET dataset in Table \ref{tab:speed}. We run all these methods on one Tesla V100 GPU three times, and report the average speed (number of sentences per second) during training and inference. Except for {\bf \textsc{LRN}} which is based on Bert-base-uncased, all the other methods are based on RoBERTa-large. Results show that {\bf \textsc{NPCRF}} (4 iterations, using the best hyper-parameters) is the fastest to model type correlations (compared with {\bf \textsc{LabelGCN}} and {\bf \textsc{LRN}}), it slows down training by 15.7\% and inference by only 13.8\%. {\bf \textsc{NPCRF}} is much faster than {\bf LITE} which is based on the cross-encoder architecture in inference\footnote{Prompt-learning based method has similar time complexity as MLC based method theoretically. We do not compare them because of differences in the PLMs and implementations.}.

\begin{table}[t]
\centering
\scalebox{0.8}{
\begin{tabular}{lcc} \toprule
\bf \textsc{Model} (sents/sec)                & \bf \textsc{Train}   & \bf \textsc{Inference}\\ \midrule
\bf \textsc{MLC-RoBERTa}    & 37.04 & 142.86 \\
\bf \textsc{MLC-RoBERTa w/ NPCRF}    & 31.25 & 123.46 \\[0.5ex] \hdashline \\[-1.5ex]
\bf \textsc{LabelGCN-RoBERTa}    & 27.78 & 90.91 \\
\bf \textsc{LRN} & 19.59 & 20.88 \\ [0.5ex] \hdashline \\[-1.5ex]
\bf \textsc{LITE} & 20.41 & 0.03 \\ \bottomrule
\end{tabular}}
\caption{Comparison of training and inference speed of different methods in Table \ref{tab:ufet}.}
\label{tab:speed}
\end{table}

\section{Conclusion and Future Work}
We propose {\bf \textsc{NPCRF}}, a method that efficiently models type correlation for ultra-fine and fine-grained entity typing, and is applicable to various entity typing backbones. In {\bf \textsc{NPCRF}}, the unary potential is formulated as the type logits of modern UFET backbones, the pairwise potentials are derived from type phrase representations that both capture prior semantic information and facilitate accelerated inference. We unfold mean-field variational inference of {\bf \textsc{NPCRF}} as a neural network for end-to-end training and inference. We find our method consistently outperforms its backbone, and reach competitive performance against very recent baselines on UFET and FET.  {\bf \textsc{NPCRF}} is efficient and require low additional computation costs. For future work, modeling higher-order label correlations, injecting prior knowledge into pairwise potentials, and extending {\bf \textsc{NPCRF}} to other tasks are worth exploring.
\section*{Limitations}
As shown in the experiments, the main limitation of {\bf \textsc{NPCRF}} is that it has less positive effect on tasks that do not require understanding type correlation (e.g., tasks with small label sets and a low number of gold labels per instance), and on models that already model label semantics quite well (e.g., prompt-based methods). Another limitation of {\bf \textsc{NPCRF}} is that, although it can be combined with many backbones, there still exist some backbones that cannot directly use {\bf \textsc{NPCRF}}, such as models that generate types one by one in an auto-regressive way (e.g., LRN) and models that cannot efficiently compute label logits (e.g., LITE).
\section{Acknowledgement}
This work was supported by the National Natural Science Foundation of China (61976139) and by Alibaba Group through Alibaba Innovative Research Program. 

\bibliography{main}

\begin{thebibliography}{44}
\expandafter\ifx\csname natexlab\endcsname\relax\def\natexlab#1{#1}\fi

\bibitem[{Arnab et~al.(2016)Arnab, Jayasumana, Zheng, and Torr}]{pcrf2}
Anurag Arnab, Sadeep Jayasumana, Shuai Zheng, and Philip~HS Torr. 2016.
\newblock Higher order conditional random fields in deep neural networks.
\newblock In \emph{European conference on computer vision}, pages 524--540.
  Springer.

\bibitem[{Chandra and Kokkinos(2016)}]{pcrf1}
Siddhartha Chandra and Iasonas Kokkinos. 2016.
\newblock Fast, exact and multi-scale inference for semantic image segmentation
  with deep gaussian crfs.
\newblock In \emph{European conference on computer vision}, pages 402--418.
  Springer.

\bibitem[{Chen et~al.(2022)Chen, Cheng, Jiang, Liu, Zhang, Shi, and
  Xu}]{chen-etal-2022-learning-sibling}
Yi~Chen, Jiayang Cheng, Haiyun Jiang, Lemao Liu, Haisong Zhang, Shuming Shi,
  and Ruifeng Xu. 2022.
\newblock \href {https://aclanthology.org/2022.acl-long.147} {Learning from
  sibling mentions with scalable graph inference in fine-grained entity
  typing}.
\newblock In \emph{Proceedings of the 60th Annual Meeting of the Association
  for Computational Linguistics (Volume 1: Long Papers)}, pages 2076--2087,
  Dublin, Ireland. Association for Computational Linguistics.

\bibitem[{Choi et~al.(2018)Choi, Levy, Choi, and Zettlemoyer}]{ufet}
Eunsol Choi, Omer Levy, Yejin Choi, and Zettlemoyer. 2018.
\newblock Ultra-fine entity typing.
\newblock In \emph{Proceedings of the ACL}. Association for Computational
  Linguistics.

\bibitem[{Coucke et~al.(2018)Coucke, Saade, Ball, Bluche, Caulier, Leroy,
  Doumouro, Gisselbrecht, Caltagirone, Lavril et~al.}]{coucke2018snips}
Alice Coucke, Alaa Saade, Adrien Ball, Th{\'e}odore Bluche, Alexandre Caulier,
  David Leroy, Cl{\'e}ment Doumouro, Thibault Gisselbrecht, Francesco
  Caltagirone, Thibaut Lavril, et~al. 2018.
\newblock Snips voice platform: an embedded spoken language understanding
  system for private-by-design voice interfaces.
\newblock \emph{arXiv preprint arXiv:1805.10190}.

\bibitem[{Dai et~al.(2021)Dai, Song, and Wang}]{mlmet}
Hongliang Dai, Yangqiu Song, and Haixun Wang. 2021.
\newblock \href {https://doi.org/10.18653/v1/2021.acl-long.141} {Ultra-fine
  entity typing with weak supervision from a masked language model}.
\newblock In \emph{Proceedings of the 59th Annual Meeting of the Association
  for Computational Linguistics and the 11th International Joint Conference on
  Natural Language Processing (Volume 1: Long Papers)}, pages 1790--1799,
  Online. Association for Computational Linguistics.

\bibitem[{Devlin et~al.(2018)Devlin, Chang, Lee, and
  Toutanova}]{devlin2018bert}
Jacob Devlin, Ming-Wei Chang, Kenton Lee, and Kristina Toutanova. 2018.
\newblock Bert: Pre-training of deep bidirectional transformers for language
  understanding.
\newblock \emph{arXiv preprint arXiv:1810.04805}.

\bibitem[{Ding et~al.(2021{\natexlab{a}})Ding, Chen, Han, Xu, Xie, Zheng, Liu,
  Li, and Kim}]{ding2021prompt}
Ning Ding, Yulin Chen, Xu~Han, Guangwei Xu, Pengjun Xie, Hai-Tao Zheng, Zhiyuan
  Liu, Juanzi Li, and Hong-Gee Kim. 2021{\natexlab{a}}.
\newblock Prompt-learning for fine-grained entity typing.
\newblock \emph{arXiv preprint arXiv:2108.10604}.

\bibitem[{Ding et~al.(2021{\natexlab{b}})Ding, Xu, Chen, Wang, Han, Xie, Zheng,
  and Liu}]{ding-etal-2021-nerd}
Ning Ding, Guangwei Xu, Yulin Chen, Xiaobin Wang, Xu~Han, Pengjun Xie, Haitao
  Zheng, and Zhiyuan Liu. 2021{\natexlab{b}}.
\newblock \href {https://doi.org/10.18653/v1/2021.acl-long.248} {Few-{NERD}: A
  few-shot named entity recognition dataset}.
\newblock In \emph{Proceedings of the 59th Annual Meeting of the Association
  for Computational Linguistics and the 11th International Joint Conference on
  Natural Language Processing (Volume 1: Long Papers)}, pages 3198--3213,
  Online. Association for Computational Linguistics.

\bibitem[{Fukushima and Miyake(1982)}]{cnn}
Kunihiko Fukushima and Sei Miyake. 1982.
\newblock Neocognitron: A self-organizing neural network model for a mechanism
  of visual pattern recognition.
\newblock In \emph{Competition and cooperation in neural nets}, pages 267--285.
  Springer.

\bibitem[{Ghamrawi and McCallum(2005)}]{pcrf2005}
Nadia Ghamrawi and Andrew McCallum. 2005.
\newblock \href {https://doi.org/10.1145/1099554.1099591} {Collective
  multi-label classification}.
\newblock In \emph{Proceedings of the 14th ACM International Conference on
  Information and Knowledge Management}, CIKM '05, page 195–200, New York,
  NY, USA. Association for Computing Machinery.

\bibitem[{Gillick et~al.(2014)Gillick, Lazic, Ganchev, Kirchner, and
  Huynh}]{ontonotes}
Dan Gillick, Nevena Lazic, Kuzman Ganchev, Jesse Kirchner, and David Huynh.
  2014.
\newblock Context-dependent fine-grained entity type tagging.
\newblock \emph{arXiv preprint arXiv:1412.1820}.

\bibitem[{Hovy et~al.(2006)Hovy, Marcus, Palmer, Ramshaw, and
  Weischedel}]{06ontonotes}
Eduard Hovy, Mitchell Marcus, Martha Palmer, Lance Ramshaw, and Ralph
  Weischedel. 2006.
\newblock \href {https://aclanthology.org/N06-2015} {{O}nto{N}otes: The 90{\%}
  solution}.
\newblock In \emph{Proceedings of the Human Language Technology Conference of
  the {NAACL}, Companion Volume: Short Papers}, pages 57--60, New York City,
  USA. Association for Computational Linguistics.

\bibitem[{Hu et~al.(2020)Hu, Jiang, Bach, Wang, Huang, Huang, and
  Tu}]{hu-etal-2020-investigation}
Zechuan Hu, Yong Jiang, Nguyen Bach, Tao Wang, Zhongqiang Huang, Fei Huang, and
  Kewei Tu. 2020.
\newblock \href {https://doi.org/10.18653/v1/2020.findings-emnlp.236} {An
  investigation of potential function designs for neural {CRF}}.
\newblock In \emph{Findings of the Association for Computational Linguistics:
  EMNLP 2020}, pages 2600--2609, Online. Association for Computational
  Linguistics.

\bibitem[{Jin et~al.(2019)Jin, Hou, Li, and Dong}]{jin-etal-2019-fine}
Hailong Jin, Lei Hou, Juanzi Li, and Tiansi Dong. 2019.
\newblock \href {https://doi.org/10.18653/v1/D19-1502} {Fine-grained entity
  typing via hierarchical multi graph convolutional networks}.
\newblock In \emph{Proceedings of the 2019 Conference on Empirical Methods in
  Natural Language Processing and the 9th International Joint Conference on
  Natural Language Processing (EMNLP-IJCNLP)}, pages 4969--4978, Hong Kong,
  China. Association for Computational Linguistics.

\bibitem[{Joshi et~al.(2017)Joshi, Choi, Weld, and
  Zettlemoyer}]{joshi2017triviaqa}
Mandar Joshi, Eunsol Choi, Daniel~S Weld, and Luke Zettlemoyer. 2017.
\newblock Triviaqa: A large scale distantly supervised challenge dataset for
  reading comprehension.
\newblock \emph{arXiv preprint arXiv:1705.03551}.

\bibitem[{Koller and Friedman(2009)}]{pgm}
Daphne Koller and Nir Friedman. 2009.
\newblock \emph{Probabilistic graphical models: principles and techniques}.
\newblock MIT press.

\bibitem[{L{\^e}-Huu and Alahari(2021{\natexlab{a}})}]{fw2021}
{\DJ}~Khu{\^e} L{\^e}-Huu and Karteek Alahari. 2021{\natexlab{a}}.
\newblock Regularized frank-wolfe for dense crfs: Generalizing mean field and
  beyond.
\newblock \emph{Advances in Neural Information Processing Systems}, 34.

\bibitem[{L{\^e}-Huu and Alahari(2021{\natexlab{b}})}]{le2021regularized}
{\DJ}~Khu{\^e} L{\^e}-Huu and Karteek Alahari. 2021{\natexlab{b}}.
\newblock Regularized frank-wolfe for dense crfs: Generalizing mean field and
  beyond.
\newblock \emph{Advances in Neural Information Processing Systems}, 34.

\bibitem[{Li et~al.(2022)Li, Yin, and Chen}]{lite}
Bangzheng Li, Wenpeng Yin, and Muhao Chen. 2022.
\newblock Ultra-fine entity typing with indirect supervision from natural
  language inference.
\newblock \emph{arXiv preprint arXiv:2202.06167}.

\bibitem[{Ling and Weld(2012)}]{figer}
Xiao Ling and Daniel~S Weld. 2012.
\newblock Fine-grained entity recognition.
\newblock In \emph{Twenty-Sixth AAAI Conference on Artificial Intelligence}.

\bibitem[{Liu et~al.(2021)Liu, Lin, Xiao, Han, Sun, and
  Wu}]{liu-etal-2021-fine}
Qing Liu, Hongyu Lin, Xinyan Xiao, Xianpei Han, Le~Sun, and Hua Wu. 2021.
\newblock \href {https://doi.org/10.18653/v1/2021.emnlp-main.378} {Fine-grained
  entity typing via label reasoning}.
\newblock In \emph{Proceedings of the 2021 Conference on Empirical Methods in
  Natural Language Processing}, pages 4611--4622, Online and Punta Cana,
  Dominican Republic. Association for Computational Linguistics.

\bibitem[{Liu et~al.(2019)Liu, Ott, Goyal, Du, Joshi, Chen, Levy, Lewis,
  Zettlemoyer, and Stoyanov}]{liu2019roberta}
Yinhan Liu, Myle Ott, Naman Goyal, Jingfei Du, Mandar Joshi, Danqi Chen, Omer
  Levy, Mike Lewis, Luke Zettlemoyer, and Veselin Stoyanov. 2019.
\newblock Roberta: A robustly optimized bert pretraining approach.
\newblock \emph{arXiv preprint arXiv:1907.11692}.

\bibitem[{L{\'o}pez and Strube(2020)}]{lopez-strube-2020-fully}
Federico L{\'o}pez and Michael Strube. 2020.
\newblock \href {https://doi.org/10.18653/v1/2020.findings-emnlp.42} {A fully
  hyperbolic neural model for hierarchical multi-class classification}.
\newblock In \emph{Findings of the Association for Computational Linguistics:
  EMNLP 2020}, pages 460--475, Online. Association for Computational
  Linguistics.

\bibitem[{Loshchilov and Hutter(2018)}]{loshchilov2018fixing}
Ilya Loshchilov and Frank Hutter. 2018.
\newblock Fixing weight decay regularization in adam.

\bibitem[{Mueller et~al.(2022)Mueller, Krone, Romeo, Mansour, Mansimov, Zhang,
  and Roth}]{mueller2022label}
Aaron Mueller, Jason Krone, Salvatore Romeo, Saab Mansour, Elman Mansimov,
  Yi~Zhang, and Dan Roth. 2022.
\newblock Label semantic aware pre-training for few-shot text classification.
\newblock \emph{arXiv preprint arXiv:2204.07128}.

\bibitem[{Onoe et~al.(2021)Onoe, Boratko, McCallum, and
  Durrett}]{onoe-etal-2021-modeling}
Yasumasa Onoe, Michael Boratko, Andrew McCallum, and Greg Durrett. 2021.
\newblock \href {https://doi.org/10.18653/v1/2021.acl-long.160} {Modeling
  fine-grained entity types with box embeddings}.
\newblock In \emph{Proceedings of the 59th Annual Meeting of the Association
  for Computational Linguistics and the 11th International Joint Conference on
  Natural Language Processing (Volume 1: Long Papers)}, pages 2051--2064,
  Online. Association for Computational Linguistics.

\bibitem[{Onoe and Durrett(2019)}]{onoe-durrett-2019-learning}
Yasumasa Onoe and Greg Durrett. 2019.
\newblock \href {https://doi.org/10.18653/v1/N19-1250} {Learning to denoise
  distantly-labeled data for entity typing}.
\newblock In \emph{Proceedings of the 2019 Conference of the North {A}merican
  Chapter of the Association for Computational Linguistics: Human Language
  Technologies, Volume 1 (Long and Short Papers)}, pages 2407--2417,
  Minneapolis, Minnesota. Association for Computational Linguistics.

\bibitem[{Pan et~al.(2022)Pan, Wei, and Zhu}]{dfet}
Weiran Pan, Wei Wei, and Feida Zhu. 2022.
\newblock Automatic noisy label correction for fine-grained entity typing.
\newblock \emph{arXiv preprint arXiv:2205.03011}.

\bibitem[{Pennington et~al.(2014)Pennington, Socher, and Manning}]{glove}
Jeffrey Pennington, Richard Socher, and Christopher Manning. 2014.
\newblock \href {https://doi.org/10.3115/v1/D14-1162} {{G}lo{V}e: Global
  vectors for word representation}.
\newblock In \emph{Proceedings of the 2014 Conference on Empirical Methods in
  Natural Language Processing ({EMNLP})}, pages 1532--1543, Doha, Qatar.
  Association for Computational Linguistics.

\bibitem[{Peters et~al.(2018)Peters, Neumann, Iyyer, Gardner, Clark, Lee, and
  Zettlemoyer}]{elmo}
Matthew~E. Peters, Mark Neumann, Mohit Iyyer, Matt Gardner, Christopher Clark,
  Kenton Lee, and Luke Zettlemoyer. 2018.
\newblock \href {https://doi.org/10.18653/v1/N18-1202} {Deep contextualized
  word representations}.
\newblock In \emph{Proceedings of the 2018 Conference of the North {A}merican
  Chapter of the Association for Computational Linguistics: Human Language
  Technologies, Volume 1 (Long Papers)}, pages 2227--2237, New Orleans,
  Louisiana. Association for Computational Linguistics.

\bibitem[{Ren et~al.(2016)Ren, He, Qu, Voss, Ji, and Han}]{ren2016label}
Xiang Ren, Wenqi He, Meng Qu, Clare~R Voss, Heng Ji, and Jiawei Han. 2016.
\newblock Label noise reduction in entity typing by heterogeneous partial-label
  embedding.
\newblock In \emph{Proceedings of the 22nd ACM SIGKDD international conference
  on Knowledge discovery and data mining}, pages 1825--1834.

\bibitem[{Shen et~al.(2017)Shen, Gan, Yan, and Zeng}]{pcrf3}
Falong Shen, Rui Gan, Shuicheng Yan, and Gang Zeng. 2017.
\newblock Semantic segmentation via structured patch prediction, context crf
  and guidance crf.
\newblock In \emph{Proceedings of the IEEE Conference on Computer Vision and
  Pattern Recognition}, pages 1953--1961.

\bibitem[{Tjong Kim~Sang and De~Meulder(2003)}]{conll03}
Erik~F. Tjong Kim~Sang and Fien De~Meulder. 2003.
\newblock \href {https://aclanthology.org/W03-0419} {Introduction to the
  {C}o{NLL}-2003 shared task: Language-independent named entity recognition}.
\newblock In \emph{Proceedings of the Seventh Conference on Natural Language
  Learning at {HLT}-{NAACL} 2003}, pages 142--147.

\bibitem[{Wainwright et~al.(2008)Wainwright, Jordan
  et~al.}]{wainwright2008graphical}
Martin~J Wainwright, Michael~I Jordan, et~al. 2008.
\newblock Graphical models, exponential families, and variational inference.
\newblock \emph{Foundations and Trends{\textregistered} in Machine Learning},
  1(1--2):1--305.

\bibitem[{Wang et~al.(2017)Wang, Li, Pavlu, and Aslam}]{pcrfT}
Bingyu Wang, Cheng Li, Virgil Pavlu, and Javed Aslam. 2017.
\newblock Regularizing model complexity and label structure for multi-label
  text classification.
\newblock \emph{arXiv preprint arXiv:1705.00740}.

\bibitem[{Weischedel and Brunstein(2005)}]{bbn}
Ralph Weischedel and Ada Brunstein. 2005.
\newblock Bbn pronoun coreference and entity type corpus.
\newblock Technical Report LDC2005T33, inguistic Data Consortium.

\bibitem[{Williams et~al.(2018)Williams, Nangia, and Bowman}]{mnli}
Adina Williams, Nikita Nangia, and Samuel Bowman. 2018.
\newblock \href {https://doi.org/10.18653/v1/N18-1101} {A broad-coverage
  challenge corpus for sentence understanding through inference}.
\newblock In \emph{Proceedings of the 2018 Conference of the North {A}merican
  Chapter of the Association for Computational Linguistics: Human Language
  Technologies, Volume 1 (Long Papers)}, pages 1112--1122, New Orleans,
  Louisiana. Association for Computational Linguistics.

\bibitem[{Wu et~al.(2019)Wu, Zhang, Mao, Guo, and Huai}]{wu2019modeling}
Junshuang Wu, Richong Zhang, Yongyi Mao, Hongyu Guo, and Jinpeng Huai. 2019.
\newblock Modeling noisy hierarchical types in fine-grained entity typing: A
  content-based weighting approach.
\newblock In \emph{IJCAI}, pages 5264--5270.

\bibitem[{Xiong et~al.(2019)Xiong, Wu, Lei, Yu, Chang, Guo, and
  Wang}]{xiong-etal-2019-imposing}
Wenhan Xiong, Jiawei Wu, Deren Lei, Mo~Yu, Shiyu Chang, Xiaoxiao Guo, and
  William~Yang Wang. 2019.
\newblock \href {https://doi.org/10.18653/v1/N19-1084} {Imposing
  label-relational inductive bias for extremely fine-grained entity typing}.
\newblock In \emph{Proceedings of the 2019 Conference of the North {A}merican
  Chapter of the Association for Computational Linguistics: Human Language
  Technologies, Volume 1 (Long and Short Papers)}, pages 773--784, Minneapolis,
  Minnesota. Association for Computational Linguistics.

\bibitem[{Xu and Barbosa(2018)}]{xu-barbosa-2018-neural}
Peng Xu and Denilson Barbosa. 2018.
\newblock \href {https://doi.org/10.18653/v1/N18-1002} {Neural fine-grained
  entity type classification with hierarchy-aware loss}.
\newblock In \emph{Proceedings of the 2018 Conference of the North {A}merican
  Chapter of the Association for Computational Linguistics: Human Language
  Technologies, Volume 1 (Long Papers)}, pages 16--25, New Orleans, Louisiana.
  Association for Computational Linguistics.

\bibitem[{Yang and Zhou(2010)}]{2phaseNER}
Li~Yang and Yanhong Zhou. 2010.
\newblock \href {https://doi.org/10.1109/BICTA.2010.5645108} {Two-phase
  biomedical named entity recognition based on semi-crfs}.
\newblock In \emph{2010 IEEE Fifth International Conference on Bio-Inspired
  Computing: Theories and Applications (BIC-TA)}, pages 1061--1065.

\bibitem[{Yavuz et~al.(2016)Yavuz, Gur, Su, Srivatsa, and
  Yan}]{yavuz-etal-2016-improving}
Semih Yavuz, Izzeddin Gur, Yu~Su, Mudhakar Srivatsa, and Xifeng Yan. 2016.
\newblock \href {https://doi.org/10.18653/v1/D16-1015} {Improving semantic
  parsing via answer type inference}.
\newblock In \emph{Proceedings of the 2016 Conference on Empirical Methods in
  Natural Language Processing}, pages 149--159, Austin, Texas. Association for
  Computational Linguistics.

\bibitem[{Zheng et~al.(2015)Zheng, Jayasumana, Romera-Paredes, Vineet, Su, Du,
  Huang, and Torr}]{crfrnn}
Shuai Zheng, Sadeep Jayasumana, Bernardino Romera-Paredes, Vibhav Vineet,
  Zhizhong Su, Dalong Du, Chang Huang, and Philip~HS Torr. 2015.
\newblock Conditional random fields as recurrent neural networks.
\newblock In \emph{Proceedings of the IEEE international conference on computer
  vision}, pages 1529--1537.

\end{thebibliography}
\bibliographystyle{acl_natbib}

\appendix

\section{Example Appendix}
\label{sec:appendix}
 
 \subsection{Pairwise Potentials}
 \label{app:potential}
\begin{figure*}[h]
     \centering
     \scalebox{0.4}{
     \includegraphics{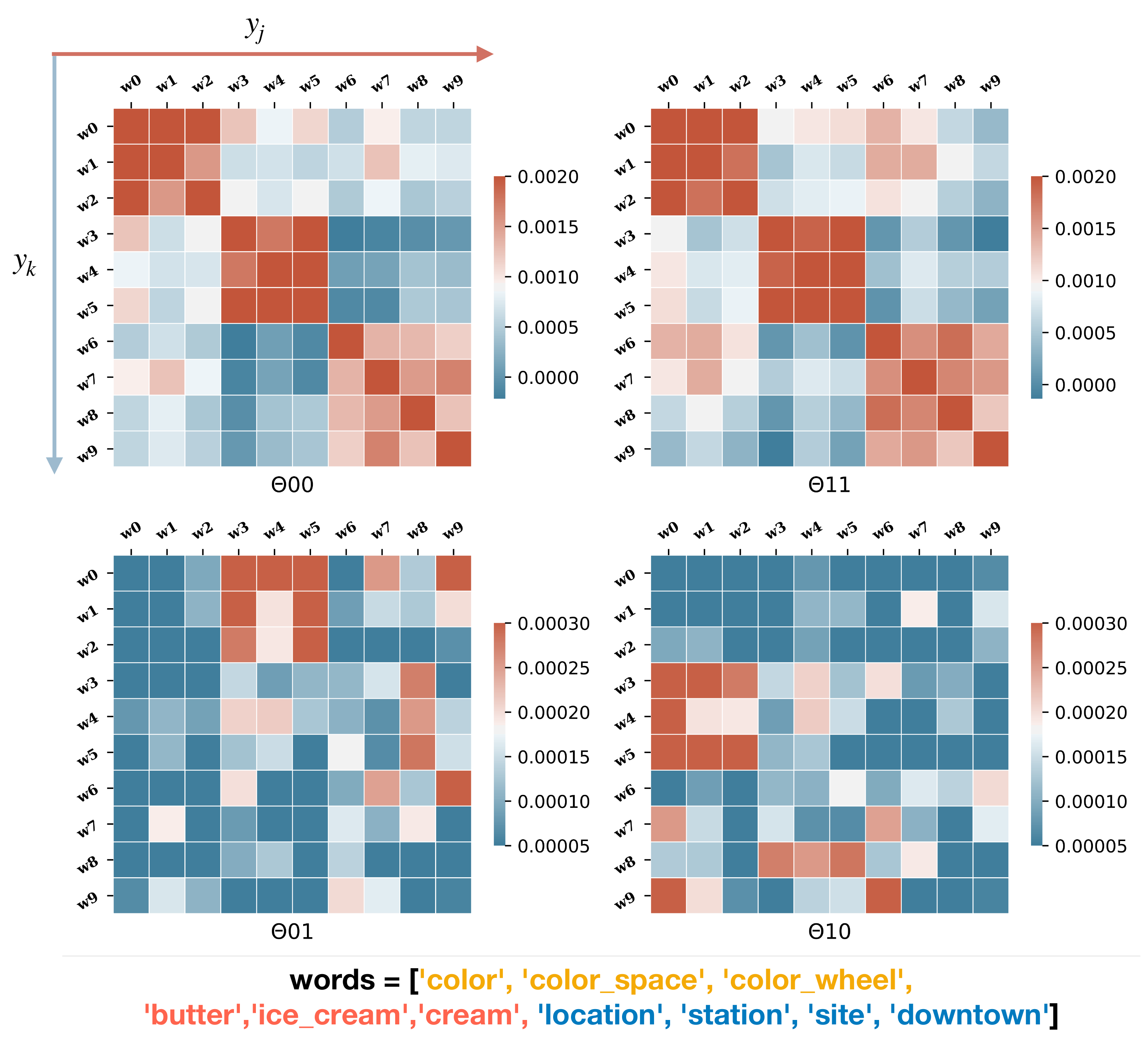}}
     \caption{Visualization of pairwise potentials. The horizontal axis is for $y_k$, and the vertical axis is for $y_j$, for example, $\theta_p(\text{location}=0, \text{downtown}=1) = \Theta_{01}[6,9]$.}
     \label{fig:pairwise}
 \end{figure*}

We recover the learned log potentials $\Theta_{00},\Theta_{11}, \Theta_{01}, \Theta_{10}$ using Eq. \ref{eq:2}. For visualization, we pick three sets of types, types in the same set are relevant, and types between different set are irrelevant, and draw the heatmap of log potentials of them in Fig. \ref{fig:pairwise}. The figure shows that: (1) the learned potentials obey the intrinsic properties, $\Theta_{00}, \Theta_{11}$ are symmetric and different from each other, $Theta_{01} = \Theta_{10}^T$ and they are asymmetric. (2) $\Theta_{00}, \Theta_{11}$ do encode different levels of similarity, the scores representing the co-absence and co-occurrence of types in the same set are higher than types in different set. (3) The $\Theta_{01}$ and $ \Theta_{01}$ obey the intrinsic properties and cells with high scores usually represent the co-exclusion of types from different sets. In general, $ \Theta_{00}$ and $ \Theta_{11}$ are less interpretable compared to $\Theta_{00}$ and $ \Theta_{11}$.

\end{document}